# A Mathematical Theory of Agency and Intelligence

Wael Hafez
Semarx Research LLC
Alexandria, VA, USA
w.hafez@semarx.com

Chenan Wei
University of Massachusetts
Amherst, MA, USA
weichenan0@gmail.com

Rodrigo Pena
Stiles-Nicholson Brain Institute,
Jupiter, FL, USA
rfdop20@gmail.com

Amir Nazeri
Semarx Research LLC
Alexandria, VA, USA
amir.nazeri@semarx.com

Cameron Reid
Semarx Research LLC
Alexandria, VA, USA
cameron.reid@semarx.com

**Abstract.** To operate reliably under changing conditions, complex systems require feedback on how effectively they use resources — not just whether objectives are met. Current AI systems process vast information to produce sophisticated predictions, yet predictions can appear successful while the underlying interaction with the environment degrades. What is missing is a principled measure of how much of the total information a system deploys is actually shared between its observations, actions, and outcomes. We prove this shared fraction — which we term bi-predictability (P) — is intrinsic to any interaction, derivable from first principles, and strictly bounded: P can reach unity in quantum systems, P ≤ 0.5 in classical systems, and lower once agency (action selection) is introduced. We confirm these bounds in a physical system (double pendulum), reinforcement learning agents, and multi-turn LLM conversations. These results distinguish agency from intelligence: agency is the capacity to act on predictions, whereas intelligence additionally requires learning from interaction, self-monitoring of its learning effectiveness, and adapting the scope of observations, actions, and outcomes to restore effective learning. By this definition, current AI systems achieve agency but not intelligence. Inspired by thalamocortical regulation in biological systems, we demonstrate a feedback architecture that monitors P in real time — establishing a prerequisite for adaptive, resilient AI.

## 1 Introduction

Modern AI excels in perception, control, and language through deep learning and extensive training (LeCun, Bengio & Hinton, 2015; Brown et al., 2020; Bommasani et al., 2021). However, general-purpose models still face reliability challenges under distribution shifts and unanticipated operating conditions (NIST, 2023; NIST, 2024; Bengio et al., 2025; O'Brien et al., 2025). Current strategies triangulate reliability by monitoring benchmark outcomes, quantifying uncertainty, and detecting input drift (Koh et al., 2021; Liang et al., 2022; Xia et al., 2025; Greco et al., 2024), while information-theoretic and causal measures quantify aspects of agent–environment interaction and control (Klyubin, Polani & Nehaniv, 2005; Salge, Glackin & Polani, 2014; Massey, 1990; Schreiber, 2000; Zhang et al., 2021; Seitzer et al., 2021; Deng et al., 2023).

Three limitations persist: monitoring often isolates fragments rather than the full observation–action–outcome loop (Sutton & Barto, 2018; Pfeifer & Bongard, 2006); feedback is used reactively (offline selection or online alarms followed by intervention) rather than as a continuous regulatory variable (NIST, 2024; Shankar et al., 2022); and signals remain task- and domain-specific, lacking a common scale. Cybernetics predicts that reliable closed-loop operation requires continuous feedback and is constrained by informational requirements on regulation and modeling (Wiener, 1948; Doyle, Francis & Tannenbaum, 1992; Åström & Murray, 2008; Ashby, 1956; Conant & Ashby, 1970).

We therefore use information as a universal currency on the full loop, defining bi-predictability, $P$, as the shared fraction of information across observations, actions, and outcomes relative to the loop's total informational budget. We prove regime-dependent bounds — unity attainable in quantum interactions, $P \leq 0.5$ classically, and lower with

agency — and use them to distinguish agency from intelligence: agency is the capacity to act on predictions, whereas intelligence additionally requires learning from interaction — that is, building predictive relationships between observations, actions, and outcomes — and self-monitoring of whether those predictions remain effective as conditions change. When they do not, intelligence demands adapting the scope of observations, actions, and outcomes to restore effective learning. Across physical, agentic, and linguistic systems, we confirm the predicted signatures of these bounds and show that $P$ can be monitored online from interaction data. Finally, we introduce an auxiliary feedback architecture (IDT), inspired by thalamocortical regulation, that monitors $P$ in real time, establishing a prerequisite signal for reliable closed-loop operation under change (Sherman & Guillery, 2006; Halassa & Kastner, 2017).

## 2 Bi-predictability ($P$)

### 2.1 Definition and Classical Bounds

We introduce a formal information-theoretic framework for quantifying how tightly two interacting entities constrain one another through their joint dynamics, independent of the absolute amount of information present or exchanged. Rather than asking how much information flows, the framework asks how much of the available uncertainty is shared—how mutually predictive the interaction is at the chosen level of description. Let $S$ and $S'$, denote successive states of the coupled system–environment interaction. We define Predictive Coherence, $P$ as:

$$P = \frac{MI(S;S')}{H(S) + H(S')} \qquad (1)$$

$P$ measures the ratio of shared information to total information—not volume, but efficiency. $P = 1/2$ corresponds to ideal closed-loop interaction where states fully determine one another; $P = 0$ corresponds to successive states being statistically independent of one another.

From first principles, Bi-predictability admits regime bounds. Under classical (Shannon) information, we obtain:

$$0 \leq P \leq \frac{1}{2} \qquad (2)$$

This bound is structural:

$$MI(S;S') \leq \min\big(H(S), H(S')\big) \qquad (3)$$

so shared information cannot exceed half of the total entropy capacity $H(S) + H(S')$ under our definition. In the quantum setting, maximally nonseparable correlations can saturate the analogous construction (as highlighted by Bell-type phenomena, (Bell, 1964)), but the transition to classical definiteness—via measurement and decoherence—removes the correlations that permit $P$ to approach unity.

### 2.2 Extension to Active Systems and Agency

The framework above applies to interacting entities in general. Many systems of interest, however—biological organisms and artificial agents—are active**:** they do not merely respond, but intervene. This introduces a natural asymmetry: one side maintains internal state and selects actions that influence what happens next.

We capture this by introducing an action variable $A$. Interaction is represented as $(S, A) \rightarrow S'$, where $S$is the agent's internal state (the information it uses to act), $A$is its chosen intervention, and $S'$is the resulting next state after the environment responds. Bi-predictability generalizes to:

$$P = \frac{MI(S, A; S')}{H(S) + H(A) + H(S')} \qquad (4)$$

Under classical (Shannon) information, the same ceiling of $1/2$ applies in principle; in practice, introducing $A$ makes this ceiling unattainable. Intuitively, action adds internal degrees of freedom that must be maintained while remaining coupled to the environment. Agents therefore trade maximal predictability for the ability to act.

Introducing $A$also makes predictability directional (Fig. 1). We define:

- Forward predictive uncertainty $H_f = H(S' \mid S, A)$: how uncertain outcomes remain given what the agent knew and did. High $H_f$indicates weak constraint of the environment's response by the agent's state and action.
- Backward predictive uncertainty $H_b = H(S, A \mid S')$: how many internal states and actions are consistent with an observed outcome. High $H_b$ indicates many distinct causes collapsing to indistinguishable consequences.

Their difference defines a predictability asymmetry**:**

$$\Delta H = H_f - H_b \qquad (5)$$

which localizes how predictability is lost—whether primarily through environmental response uncertainty (high $H_f$) or through agent-side indistinguishability (high $H_b$). This decomposition matters: two systems can exhibit similar $P$ yet fail for different reasons. $P$ measures overall coupling efficiency; $\Delta H$ reveals where the coupling breaks.

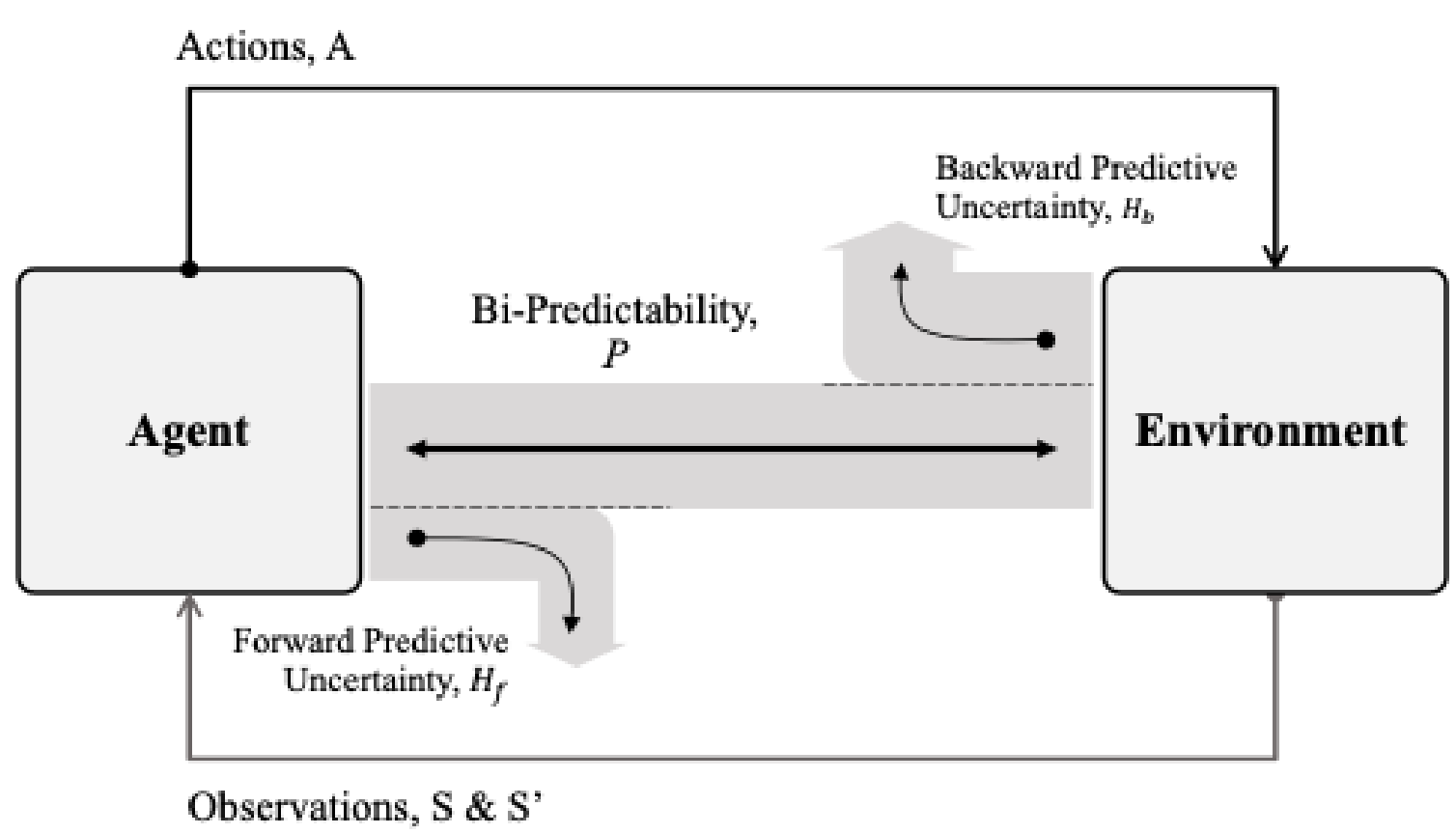

***Figure 1*** *Bi-predictability decomposes into directional predictive uncertainties. P = bi-predictability;* $H_f$ *= H(S'|S,A), forward predictive uncertainty;* $H_b$ *= H(S,A|S'), backward predictive uncertainty.*

$P$ and $\Delta H$ expose the trade-off introduced by agency: acting adds freedom, but intelligent action requires outcomes to be both controllable (low forward uncertainty) and legible (low backward ambiguity). Similar $P$ values can hide different failure modes; $\Delta H$ separates them by indicating whether predictability is lost mainly in the environment's response ($H_f$) or in the agent's own indistinguishability ($H_b$).

$P$ is not a normalization in the statistical sense—the denominator exceeds the numerator's theoretical maximum $P$ measures informational yield relative to total deployed capacity, not proximity to perfect coupling.

# 3 Physical Calibration: Double Pendulum Confirms the Classical Bound

We first test the framework on a deterministic physical system without an action channel: the double pendulum. This establishes a calibration point in which any loss of predictability arises from measurement/representation rather than decision-making.

## 3.1 P Tracks Chaos Regime Transitions

We analyzed two batches of 300 simulations spanning symmetric (equal mass) and asymmetric (unequal mass) settings.

Prediction 1: High Bi-predictability. Under deterministic dynamics with a complete state representation, $P$ should approach the classical ceiling of $1/2$. Consistent with this prediction, $P$ remains close to the bound across both batches with low variance (Table 1), indicating that successive states are strongly mutually predictive despite chaotic sensitivity.

***Table 1*** *Bi-predictability in the double pendulum under symmetric and asymmetric mass configurations. P = bi-predictability.*

| Batch | P Min | P Mean | P Max | P STD |
|---|---|---|---|---|
| **1st batch** | 0.472944657 | 0.475747126 | 0.48095266 | 0.00155264 |
| **2nd batch** | 0.475749647 | 0.472919123 | 0.4769658 | 0.00169209 |

Prediction 2: Predictive asymmetry ≈ 0. In the absence of intervention or intrinsic randomness, forward and backward predictive uncertainty should be comparable, yielding $\Delta H \approx 0$. As predicted, forward and backward uncertainties are numerically indistinguishable and $\Delta H$ is centered near zero across both batches (Table 2).

***Table 2*** *Forward and backward predictive uncertainty in the double pendulum.* $H_f = H(S'|S)$*, forward predictive uncertainty;* $H_b = H(S|S')$*, backward predictive uncertainty;* $\Delta H = H_f - H_b$*, predictive asymmetry.*

| Batch | Metric | Min | Mean | Max | STD |
|---|---|---|---|---|---|
| **1st batch** | Forward Predictive Uncertainty, $H(S'\|S)$ | 0.101762767 | 0.173011607 | 0.21491649 | 0.02207975 |
| | Backward Predictive Uncertainty $H(S\|S')$ | 0.101853985 | 0.172982039 | 0.21480427 | 0.02213793 |
| | Predictive Asymmetry $\Delta H = H(S'\|S) - H(S\|S')$ | -3.76E-04 | -6.6996E-07 | 5.15E-04 | 0.00016409 |
| **2nd batch** | Forward Predictive Uncertainty $H(S'\|S)$ | 7.62E-02 | 0.132628187 | 0.19915062 | 0.01919794 |
| | Backward Predictive Uncertainty $H(S\|S')$ | 7.63E-02 | 0.132629277 | 0.19949738 | 0.01920498 |
| | Predictive Asymmetry $\Delta H = H(S'\|S) - H(S\|S')$ | -6.25E-04 | -1.09E-06 | 0.00063149 | 0.00022369 |

Prediction 3: Chaos does not imply asymmetry. Across high-chaos regimes, $P$ remains stable and $\Delta H$ remains near zero, supporting the distinction between chaotic sensitivity and directional loss of predictability in this setting.

## 3.2 Interpretation

Together, these results establish a physical calibration: for a deterministic system without an action channel, $P$ approaches the classical ceiling and $\Delta H$ remains near zero. The small gap from the theoretical maximum is consistent with finite estimation and representation effects (for example, discretization and windowing) rather than dynamical limitations. In later agentic settings, departures from this pattern indicate that predictability is being lost through intervention and/or openness at the chosen interface.

# 4 From Agency to Intelligence: An Information Architecture

## 4.1 Conditions for Agency

A system exhibits **agency** when an action variable $A$satisfies three conditions:

- Choice: $H(A \mid S) > 0$— actions are not fully determined by the available state.
- Effect: $MI(A; S' \mid S) > 0$— actions change what happens next beyond what the state already predicts.
- Predictive asymmetry: $\mid \Delta H \mid > 0$— forward and backward predictive uncertainty differ.

Choice and effect are structural: the system can select among alternatives, and those alternatives matter. Predictive asymmetry is diagnostic: it indicates directional intervention at the $(S, A, S')$ interface.

In deterministic physical dynamics without an action channel, forward and backward uncertainty remain balanced ($\Delta H \approx 0$) at the chosen description level, even under chaos. The double pendulum provides this baseline. Introducing action typically breaks this balance: outcomes do not fully "round-trip" back to the agent's internal causes, producing a measurable asymmetry in the predictive structure.

## 4.2 Conditions for Intelligence

Agency enables intervention; intelligence manages the quality of that intervention. We define intelligence as requiring three capacities:

- Learning: increase overall interaction predictability $MI(S, A; S')$ (the numerator of $P$).
- Self-monitoring: measure and regulate $P$over time.
- Adaptation: expand or reorganize the state, action, and outcome spaces $\{S\}, \{A\}, \{S'\}$—that is, change what the system can represent, what it can do, and what outcomes it can reliably bring about.

Learning builds coupling within a fixed interface; self-monitoring evaluates coupling efficiency; adaptation reshapes the interface itself.

By this definition, current AI typically achieves agency and learning, but lacks explicit self-monitoring and adaptation. Training can increase $MI(S, A; S')$ while leaving coupling efficiency unmeasured and $\{S\}, \{A\}, \{S'\}$ fixed by designers, which is why degradation detection still relies on external evaluation rather than first-person monitoring.

## 4.3 The Information Digital Twin (IDT)

To enable self-monitoring, we propose the Coupled Agency Architecture, which pairs the agentic policy with a regulatory Information Digital Twin (IDT) (Fig. 2). Unlike standard twins that replicate physical states, the IDT models interaction statistics, functioning as a homeostatic sidecar independent of the agent's internal model. The architecture operates in three stages: (1) Metric Estimation, where the IDT computes real-time $P$ and $\Delta H$ from the $(S, A, S')$ stream; (2) Stability Control, where a '$P$ Controller' detects statistical deviations from the coherent baseline; and (3) Reflexive Modulation, where significant excursions trigger interaction information efficiency modulation. By employing signal management techniques—such as action dampening ("Hold"), input filtering, or dimensionality reduction—this mechanism resolves open-loop fragility without requiring immediate retraining. In this way, the IDT supports real-time stability and provides further actionable insights for the adaptation of $\{S\}, \{A\}, \{S'\}$, to further improve agent Bi-predictability and ultimately its decision effectiveness.

By modulating the interface rather than the model weights, the system preserves agency during perturbations that would otherwise cause catastrophic drift. This functionally mirrors the mammalian thalamocortical loop, where thalamic nuclei monitor copies of sensory and motor signals and regulate signal transmission based on signal statistics rather than semantic content. The IDT thus provides the necessary engineering blueprint for converting passive predictive metrics into active, homeostatic agency.

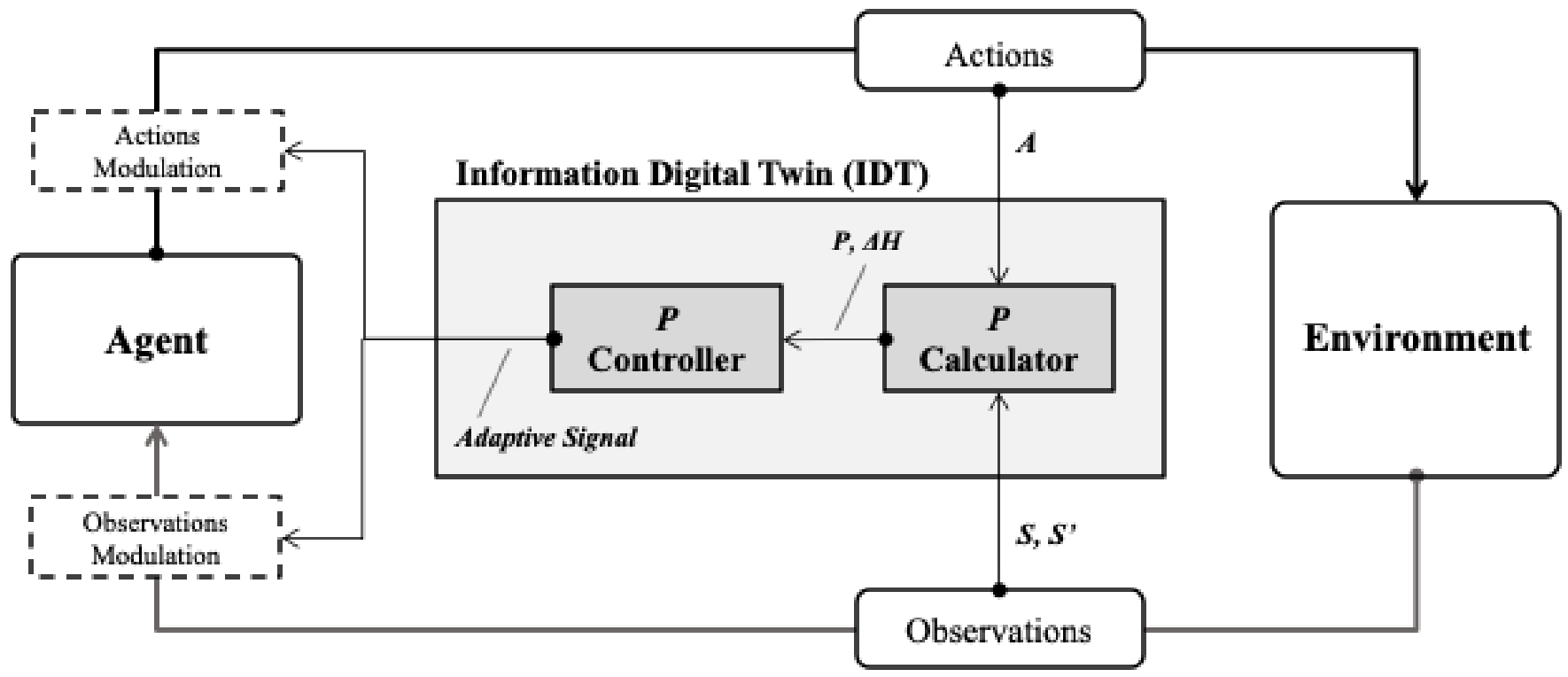

***Figure 2*** *Information Digital Twin (IDT) architecture. The IDT operates alongside the agent–environment loop, receiving copies of observations (S, S') and actions (A). The P Calculator computes bi-predictability P and predictive asymmetry ΔH from the interaction stream. The P Controller detects statistical deviations from baseline coupling. Dashed boxes indicate architecturally specified modulation pathways — observation modulation and action modulation — not experimentally validated in this work. P = bi-predictability; ΔH = predictive asymmetry ($H_f - H_b$).*

### 4.4 Differentiation from Existing Frameworks

Existing frameworks define agency through feedback and stability (Wiener, 1948; Ashby, 1956), reward optimization robustness failures (Amodei et al., 2016; D'Amour et al., 2022; Shumailov et al., 2024), or intrinsic motivation such as empowerment (Schmidhuber, 1991; Klyubin et al., 2005) and prediction error minimization (Rao & Ballard, 1999; Friston, 2010). These approaches share a limitation: they measure unidirectional influence—agent → environment (empowerment) or environment → agent (prediction error)—not bidirectional coupling.

Bi-predictability differs: $P$ measures mutual coupling; $\Delta H$ attributes degradation to environmental variability ($H_f$) or internal indistinguishability ( $H_b$). This matters for coordination—agents unpredictable in their effects become unreliable partners (Dragan et al., 2013; Hadfield-Menell et al., 2016), $H_b$ directly quantifies this failure mode.

### 4.5 Biological Precedent: Thalamocortical Regulation

Intelligence, as defined here, requires observing the ($S, A, S'$) stream. A biological precedent exists in the mammalian thalamocortical loop, where thalamic nuclei receive copies of both sensory signals ($S$) and motor commands ($A$) via branching axons (Guillery, 2005). These are copies—not modulatory inputs—positioning the thalamus as an observer of the interaction, not a controller. Thalamic circuits operate on signal statistics—gain, synchrony, bandwidth—rather than semantic content (Sherman & Usrey, 2024; Cassidy et al., 2025). This suggests biology monitors interaction structure independently of task meaning. We do not claim the thalamus implements an IDT. Rather, it provides existence proof that copy-based observation of ($S, A$) streams can coexist with effective control—an architectural principle evolution discovered independently.

## 5 Engineering Validation

We test whether current AI systems satisfy the operational conditions for agency and intelligence introduced above. This extends prior work showing that interaction information $MI(S, A; S')$ can flag behavioral anomalies in robotics and perception (Reid et al., 2025; Nazeri et al., 2025) by adding regime bounds and explicit criteria. We evaluate reinforcement-learning agents in continuous control and large language model agents in multi-turn interaction, computing $P$ and $\Delta H$ from the ($S, A, S'$) stream without access to model internals, reward shaping, or semantic content.

## 5.1 Bi-predictability for Reinforcement Learning Agents (RL)

### 5.1.1 Experimental Setup

We evaluate continuous-control agents in MuJoCo (Todorov et al., 2012), trained with SAC and PPO (Haarnoja et al., 2018; Schulman et al., 2017) on HalfCheetah. Policies are frozen during evaluation. Metrics $P$, $H_f$, $H_b$, and $\Delta H$ are computed over fixed-length sliding windows; perturbations begin mid-evaluation after a baseline period. Results aggregate across seeds (11 SAC, 10 PPO). For threshold-based detection, seeds with unstable pre-perturbation baselines were excluded from detection-rate summaries (reported explicitly below), since calibration requires a stationary baseline.

### 5.1.2 Trained Agents Converge to P ≈ 0.33

Under normal operation, Half-Cheetah exhibits $P = 0.33 \pm 0.02$ and $\Delta H = -0.56 \pm 0.22$, placing it below the classical ceiling and within the agentic regime. The negative $\Delta H$ indicates persistent asymmetry: backward ambiguity exceeds forward uncertainty, consistent with interventions that do not fully round-trip from outcomes back to internal causes. Table 3 contrasts this with the double pendulum baseline ($P \approx 0.48$, $\Delta H \approx 0$), separating physical and agentic regimes.

***Table 3*** *Bi-predictability across physical and agentic systems. P values for the double pendulum (passive), RL agents (SAC, PPO), and LLM interactions. P = bi-predictability. See main text for interpretation.*

| System | P | ΔH | Interpretation |
|---|---|---|---|
| **Double pendulum** | 0.48 | ≈ 0 | Physics: high bi-predictability, symmetric prediction |
| **HalfCheetah (baseline)** | 0.33 | -0.56 | Agency: reduced bi-predictability, asymmetric prediction |

### 5.1.3 IDT Detects Perturbations Reward Misses

We injected eight perturbation types spanning environment-side changes (e.g., forces/gravity) and agent-side degradation (e.g., observation/action noise). Across 168 perturbation trials, the IDT detected $89.3 \pm 15.1\%$ of perturbations, compared with $44.0 \pm 26.1\%$ using reward-based detection ($t = 7.95, p < 10^{-6}$). Individual components ($P$, $\Delta H$, $H_f$, $H_b$) each detect several perturbations, and their union increases coverage because the signals respond to different failure modes.

### 5.1.4 IDT Detects Degradation 4.4× Faster

IDT also detects degradation earlier. Median detection latency is 42 windows post-onset for IDT versus 184 for reward (Table 4), reflecting that reward integrates effects over many transitions whereas $P$and $\Delta H$ track coupling integrity at the transition level.

***Table 4*** *Detection latency: median number of windows from perturbation onset to first threshold exceedance. IDT = Information Digital Twin union of P, $\Delta H$, $H_f$, $H_b$.*

| Metric | Median Latency (windows) |
|---|---|
| *IDT* | 42 |
| $\boldsymbol{P}$ | 74 |
| $\boldsymbol{\Delta H}$ | 67 |
| $\boldsymbol{H_f}$ | 69 |
| $\boldsymbol{H_b}$ | 75 |
| Rewards | 184 |

### 5.1.5 Bi-Predictability Captures Silent Degradation

These results support the framework's central distinction between task performance and interaction quality. Baseline values ($P = 0.33$, $\Delta H = -0.56$) place the RL agent below the physical ceiling and show the predictive asymmetry expected in the agentic regime. The detection advantage (89% $vs$ 44% coverage; 4.4 × lower median latency) follows from what the signals measure: reward integrates outcomes over many transitions, so coupling degradation often becomes visible only after failures accumulate. By contrast, $P$ and $\Delta H$ track coupling at the transition level, so disruption is detectable immediately—even before returns degrade.

Because $P$ and $\Delta H$ respond to different failure modes, their combination increases detection coverage beyond any single component (Fig. 4). Moreover, different perturbations produce distinct response patterns across $P$, $H_f$, $H_b$, and $\Delta H$, suggesting a path toward attribution rather than a single undifferentiated alarm.

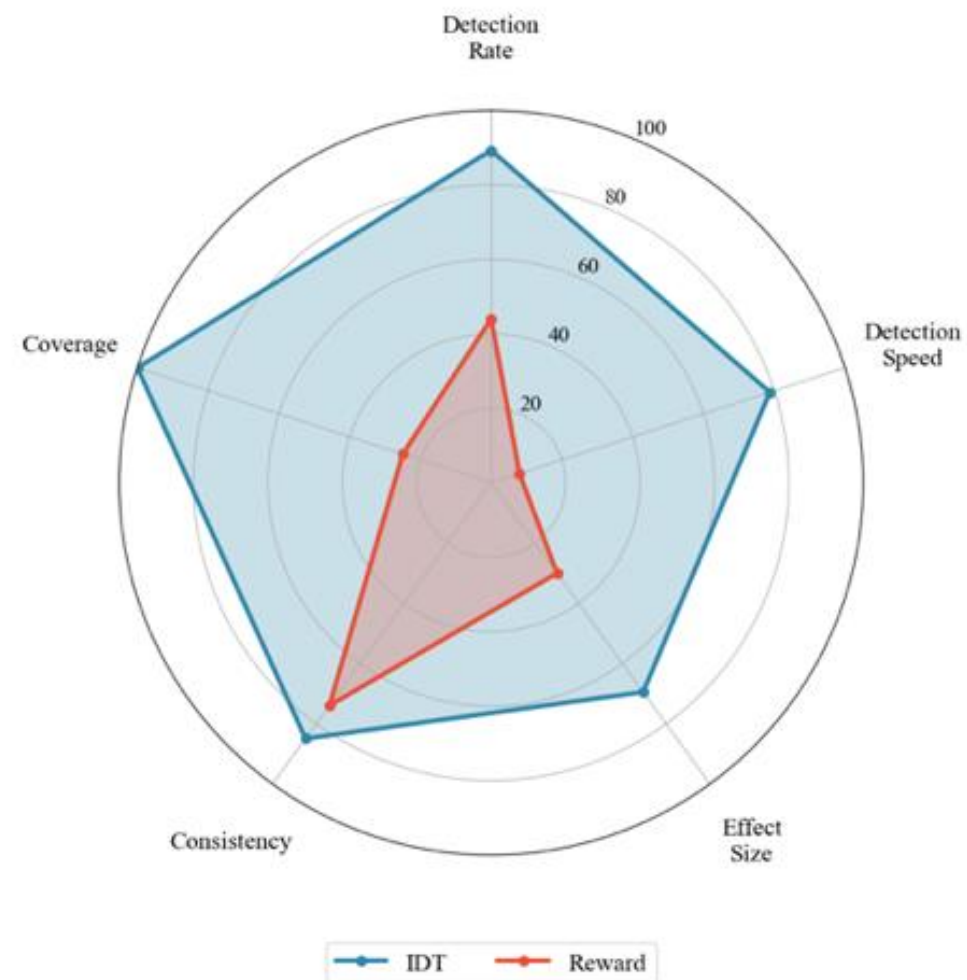

***Figure 3*** *IDT and reward-based monitoring comparison across five diagnostic dimensions. Blue: IDT (union of P, $H_f$, $H_b$, and $\Delta H$). Red: reward-based detection. Dimensions: detection rate, speed, effect size, consistency, and coverage. IDT = Information Digital Twin.*

### 5.1.6 Current RL Agents Achieve Agency but Not Intelligence

RL agents satisfy all three agency conditions: stochastic policies ensure choice ($H(A|S) > 0$), actions causally influence outcomes ($MI(A;S'|S) > 0$), and predictive asymmetry distinguishes them from passive physics ($\Delta H = -0.56 \neq 0$). They also satisfy learning—training maximizes $MI(S,A;S')$ towards cumulative reward. However, they lack self-monitoring and adaptation: no mechanism computes P from the agent's own $(S,A,S')$ stream, nor can they adjust their sensors ($S$), effectors ($A$), or deployment environment ($S'$). By our definition, current RL agents exhibit agency and learning, but not intelligence (Table 5).

***Table 5*** *Agency and intelligence conditions (RL agents). RL agents satisfy agency (choice, effect, asymmetry) and learning, but lack self-monitoring—the defining gap between current AI and intelligence. P = bi-predictability; IDT = Information Digital Twin; SAC = Soft Actor-Critic; PPO = Proximal Policy Optimization.*

| Condition | | Criterion | Evidence | Achieved |
|---|---|---|---|---|
| Agency | Choice | $H(A\|S) > 0$ | Stochastic policies (SAC, PPO) | Yes |
| | Effect | $MI(A;S'\|S) > 0$ | Actions influence outcomes | Yes |
| | Asymmetry | $\|\Delta H\| > 0$ | $\Delta H = -0.56 \pm 0.22$ | Yes |
| Intelligence | Learning | $\uparrow MI(S,A;S')$ towards objective | Trained on $(S,A,S',R)$ to maximize reward | Yes |
| | Self-monitoring | Computes $P$ from own stream | No internal $P$ computation | NO |
| | Adaptation | Adjusts $\{S\},\{A\},\{S'\}$ | Spaces fixed by designers | NO |

## 5.2 Bi-Predictability Generalizes to Language Model Interactions

### 5.2.1 Setup

To test generality beyond physical control, we evaluate Bi-predictability in multi-turn dialogue. A student model (Llama 3.1 8B) interacts for 85–200 turns with three distinct teacher models (Claude, ChatGPT, Gemini) across 34 unique test–teacher–condition combinations (4,574 turns total). Conditions varied: normal (temperature 0.7, top_k 40) allowed unrestricted generation, while constrained (temperature 0.1, top_k 10) reduced response diversity, simulating capacity degradation.

Three baseline tests examined natural conversation dynamics using prompts designed to elicit varied questioning styles. Three perturbation tests evaluated sensitivity to conversational disruptions—contradictions, topic shifts, and non-sequiturs—injected at fixed intervals after a 30-turn baseline. We map dialogue into the $(S, A, S')$ loop: $S$ is accumulated context, $A$ is the student response, and $S$' is the teacher's subsequent prompt. Metrics, $H_f$, $H_b$, and $\Delta H$ are computed from token-frequency distributions.

### 5.2.2 Structural Coupling Metrics Complement Semantic Evaluation

We compare $P$and $\Delta H$against two widely used baselines: embedding-based cosine similarity for structural consistency (Reimers & Gurevych, 2019) and LLM-as-a-judge for semantic quality (Zheng et al., 2023). Across conditions, $P$ aligns strongly with structural consistency (significant correlation in 85% of cases; Table 6) but aligns less reliably with judge scores (44% of cases). This separation indicates that $P$ primarily tracks interaction structure rather than semantic correctness—an interaction-quality signal that does not require embeddings or external evaluation models.

***Table 6*** *Relationship to structure and semantics. Across test conditions, $P$and $\Delta H$ correlate more consistently with embedding-based structural similarity than with judge-based semantic scores, indicating that Bi-predictability primarily tracks interaction structure. P = bi-predictability; ΔH = predictive asymmetry ($H_f - H_b$).*

| Metric | Correlation with Structure (Cosine Sim) | Correlation with Semantics (LLM Judge) |
|---|---|---|
| Prediction Efficiency (***P***) | **85%** (29/34 conditions) | 44% (15/34 conditions) |
| Prediction Asymmetry (**Δ*H***) | **76%** (26/34 conditions) | 47% (16/34 conditions) |

### 5.2.3 Bi-Predictability Detects LLM Drift across Teacher Models

We inject three perturbation types (contradictions, topic shifts, non-sequiturs) at fixed turn positions. Using only token statistics, $P$ and $\Delta H$ achieved 100% detection across all teacher models and perturbation types ($9/9$ trials per condition, $p < 0.001$), matching the sensitivity of semantic judges (Cosine/GPT-4) but with significantly lower computational overhead. As shown in Fig. 4, deviations exhibit consistent signatures: $P$ exhibits immediate instability at injection points—typically a sharp drop due to confusion or occasionally a spike due to fixation—while backward predictivity ($H_b$,) simultaneously increases. This confirms that structural coupling metrics are sufficient to flag semantic breakdowns without requiring heavy semantic evaluation

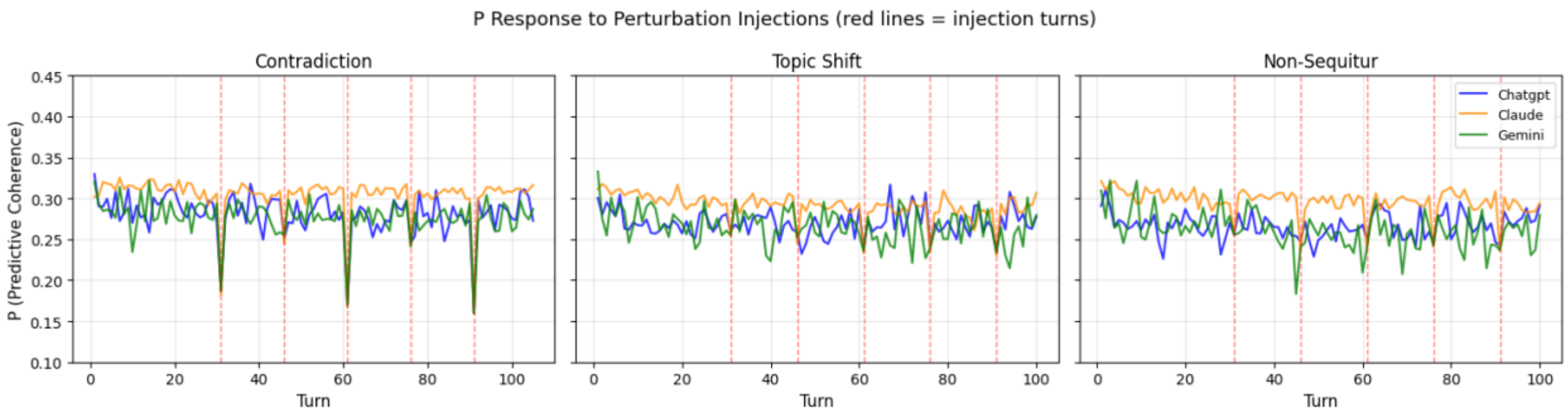

***Figure 4*** *Bi-predictability response to dialogue perturbations across three teacher models. (a) Contradiction injection, (b) topic shift injection, (c) non-sequitur injection. Each panel shows P over 100 conversational turns for three teacher models: ChatGPT (blue), Claude (orange), and Gemini (green). Red dashed vertical lines indicate perturbation injection turns. P = bi-predictability, defined as* $MI(S, A; S') / [H(S) + H(A) + H(S')]$.

### 5.2.4 LLMs Exhibit Agency but Not Intelligence

LLM agents satisfy the agency criteria at the interaction level: stochastic sampling provides choice; responses influence subsequent context; and $\Delta H$ indicates persistent predictive asymmetry. They also satisfy learning through next-token training. However, as summarized in Table 8, current LLM agents lack explicit self-monitoring and adaptation: they do not compute coupling quality, nor can they reorganize their interface in response to degradation.

The IDT fills this gap. Unlike semantic evaluators such as cosine similarity or LLM judges—which introduce significant latency and model dependencies—the IDT operates directly on raw token statistics with negligible overhead. This computational efficiency allows it to transform the passive token stream into a real-time active control signal. By surfacing bi-predictability deviations immediately, the IDT provides the necessary feedback to trigger the Coupled Agency Architecture's reflexive modulation—enabling the system to restore stability through context gating or parameter adjustment, rather than relying solely on fixed next-token probabilities.

***Table 7*** *Agency and intelligence conditions in LLM agents. LLM agents satisfy agency and learning, but lack explicit self-monitoring and adaptation under our definition. LLM = large language model; IDT = Information Digital Twin.*

| Condition | | Criterion | Evidence | Achieved |
|---|---|---|---|---|
| Agency | Choice | $H(A\|S) > 0$ | Stochastic sampling (temperature > 0) | Yes |
| | Effect | $MI(A; S'\|S) > 0$ | Responses influence subsequent context | Yes |
| | Asymmetry | $\|\Delta H\| > 0$ | $\Delta H < 0$ across all conditions | Yes |
| Intelligence | Learning | $\uparrow MI(S, A; S')$ towards objective | Trained on token sequences to predict next token | Yes |
| | Self-monitoring | Computes $P$ from own stream | No internal $P$ computation | NO |
| | Adaptation | Adjusts $\{S\}, \{A\}, \{S'\}$ | Vocabulary and generation parameters (context window, top_p, top_k, max response) fixed by designers/users | NO |

## 6 Discussion

Currently, AI development focuses on scaling the internal model (learning). Our results suggest that reliable agency requires a parallel focus on the Information Architecture: the structural capacity to regulate coupling quality. By identifying Predicative Coherence ($P$) as the order parameter of interaction, we distinguish effective agency from mere throughput. The systematic reduction of $P$ when actions are introduced reflects the informational cost of freedom; intelligence is not the elimination of this cost, but the active management of it via self-monitoring.

Within this framework, agency is the introduction of choice into the agent–environment loop, while intelligence requires learning plus explicit self-monitoring and adaptation. Actions add internal degrees of freedom that typically reduce raw predictability; managing this trade—rather than eliminating it—is the defining challenge of adaptive behavior. Both reinforcement-learning and large language model agents satisfy agency (choice, effect, asymmetry) and learning (increasing interaction predictability toward objectives). Yet neither satisfies self-monitoring nor adaptation: no current AI computes its own decision effectiveness from its own interaction stream, and state–action–outcome spaces remain designer-defined. Thus, under our operational definition, current AI exhibits agency and learning, but not intelligence.

Accordingly, there is a need for a metric that captures the "first-person" structural state of the agent, distinct from its third-person objective performance. While reward functions track external success, $P$ quantifies the agent's "grip" on the environment—the bidirectional constraint where perception reliably dictates outcomes (forward predictability) and outcomes unambiguously reveal authorship (backward predictability). In biological systems, the independent failure of these constraints corresponds to distinct breakdowns requiring distinct recoveries. High forward uncertainty ($H_f$) means the world is opaque to the agent—outcomes remain unpredictable despite action. High backward uncertainty ($H_b$) means the agent is opaque to the world—different actions produce indistinguishable outcomes, as if the environment cannot read the agent's intent. Without this differentiation, an agent knows only that $P$ dropped, not whether to adjust its predictions ($H_f$) or its legibility ($H_b$). Attribution is not diagnostic luxury—it is prerequisite for effective adaptation. Current AI systems are blind to these structural shifts; they pursue objectives even as causal coupling disintegrates. $P$ $and$ $\Delta H$ together provide the missing first-person metric: $P$ measures coupling integrity, $\Delta H$ indicates where it fails.

The metric $P = MI(S, A; S') / H_total$ operationalizes this by quantifying the fraction of total system entropy captured by the state-action-next-state coupling. Any significant deviation from baseline — regardless of direction — indicates the learned information structure no longer holds. The Information Digital Twin (IDT) monitors this coupling in real-time, supplying the regulatory layer missing from reward-based systems. By separating 'Task Performance' (the Agent) from 'Coupling Stability' (the IDT), the proposed Coupled Agency Architecture resolves the fragility of open-loop control. We identify Reflexive Modulation — the ability to gate observation and action bandwidths in response to statistical drift — as the critical mechanism for recovery. This mirrors the mammalian thalamus, which regulates signal transmission based on statistical properties rather than semantic content. While we define the information-theoretic specifications for these modulation interfaces, the specific control laws mapping predictability deviations to bandwidth adjustments remain a domain-specific engineering challenge for future work.

Collectively, these results establish that scalable intelligence depends not only on objective functions, but on explicitly engineered information coupling architectures—a structural layer that biological systems embody and current artificial systems must now adopt.

## 7 Conclusion

This paper establishes a mathematical framework for characterizing the information structure of any interaction between a system and its environment. We derive bi-predictability ($P$) — the fraction of total information shared between observations, actions, and outcomes — as an intrinsic property of interaction, prove regime-dependent bounds ($P \leq 1$ for quantum systems, $P \leq 0.5$ for classical systems, and lower once agency is introduced), and provide operational definitions that distinguish agency from intelligence. Across a physical system (double pendulum), reinforcement learning agents, and multi-turn LLM conversations, the predicted signatures of these bounds are confirmed and P is shown to be computable online from interaction data alone.

These results reframe a central challenge in AI. The field currently treats reliability as a training problem — solvable by scaling data, parameters, and compute. Our findings suggest it is fundamentally an architectural problem: current AI systems build predictive relationships but have no mechanism to monitor whether those predictions remain effective, nor to restructure what they observe or do when predictions degrade. This structural absence — not insufficient scale — is what separates agency from intelligence.

The theory establishes the requirement; the engineering program that follows is necessarily domain-specific. How a reinforcement learning agent restructures its observation and action spaces in response to declining $P$ will differ fundamentally from how an LLM manages its context and response strategy. Developing these domain-specific adaptation mechanisms — closing the loop from monitoring to modulation — is the immediate next step. The Information Digital Twin demonstrated here provides the prerequisite: a task-independent, model-agnostic, real-time signal that such adaptation requires.

More broadly, bi-predictability reveals that the tradeoff between predictive grip and freedom of action is not a design choice but a physical constraint, governed by provable bounds that tighten as systems gain autonomy. Understanding how interacting systems navigate this tradeoff — maintaining sufficient information proximity to act effectively while preserving the independence to act at all — may prove foundational not only for engineering intelligent AI but for any discipline concerned with how complex systems sustain effective interaction with their environments under change.

# 8 Method: Deriving $P$ from first Principles

## 8.1 Bi-predictability $P$ (Discrete Case): Definition and Bound

We define the general tripartite case $(S, A, S')$, which captures agentic interaction. Passive physical systems—where no action variable intervenes—are a special case with $H(A) = 0$ and $P = MI(S; S') \,/\, [H(S) + H(S')]$. System Definition. We model the agent-environment interaction as a stochastic process involving three discrete random variables: Observation $S \in \{S\}$, Action $A \in \{A\}$, and Outcome $S' \in \{S'\}$ We assume the state spaces *{S}, {A}, {S'}* are finite. The system's total information capacity, *C*, is defined as the sum of the marginal entropies:

$$C = H(S) + H(A) + H(S')$$

where $H(.)$ denotes the Shannon entropy in bits.

Metric Definition. We define Bi-predictability ($P$) as the ratio of the interaction information contributing to the prediction of the outcome to the total system capacity:

$$P = \frac{MI(S, A; S')}{C}$$

The numerator, $MI(S, A;\ S')$, quantifies the reduction in uncertainty about the outcome $S'$ given the joint knowledge of the observation S and action A. Using the chain rule for mutual information, this can be expanded as:

$$MI(S, A;\ S') = MI(S;\ S') + MI(A;\ S' \mid S)$$

This decomposition highlights that predictive power is the sum of the environment's passive predictability ($MI(S;\ S')$) and the agent's active causal contribution ($MI(A;\ S' \mid S)$).

## 8.2 Derivation of the Universal Bound.

To establish the upper bound of P, we maximize the numerator $M = MI(S, A;\ S')$ subject to the constraint *C* = *constant*. From the definition of mutual information, *M* is bounded by the entropies of the variables involved:

1. $M = H(S') - H(S' \mid S, A) \leq H(S')$
2. $M = H(S, A) - H(S, A \mid S') \leq H(S, A)$

Combining these, $M \leq min\,(H(S, A), H(S'))$. Since the joint entropy is bounded by the sum of marginal entropies, $H(S, A) \leq H(S) + H(A)$, we arrive at the weaker but operational bound:

$$M \leq (H(S) + H(A), H(S'))$$

Let $X = H(S) + H(A)\ and\ Y = H(S')$. The constraint becomes $X + Y = C$. The function $min(X, Y)$ under the constraint $X + Y = C$ is maximized when $X = Y = C/2$. Thus, the maximum possible value for the numerator is $C/2$. Substituting this into the definition of P:

$$P_max = (C/2)/C = 1/2$$

This proves that $P \leq 1/2$ is a universal bound for any classical system representable by these variables.

## 7.3 Quantum Bound

For quantum systems, Bi-predictability is defined using von Neumann entropy $S(\rho) = -Tr(\rho \log \rho)$ in place of Shannon entropy:

$$P = I(S{:}S') / [S(\rho_S) + S(\rho_S')]$$

where $I(S{:}S') = S(\rho_S) + S(\rho_S') - S(\rho_SS')$ is the quantum mutual information.

For a maximally entangled bipartite state (e.g., a Bell pair), the marginal states are maximally mixed: $S(\rho_S) = S(\rho_S') = 1$. The joint state is pure: $S(\rho_SS') = 0$. Therefore:

$$I(S{:}S') = 1 + 1 - 0 = 2$$

$$P = 2 / (1 + 1) = 1$$

This bound is unachievable classically. Classical correlations satisfy $MI(X;Y) \leq min(H(X), H(Y))$, which implies $P \leq 1/2$ (Cover & Thomas, 2006). Entanglement enables correlations captured by quantum mutual information that have no classical counterpart under Shannon information, allowing the quantum analogue of $P$ to reach unity for maximally entangled pairs (Nielsen & Chuang, 2010). The transition from quantum ($P \leq 1$) to classical ($P \leq 0.5$) reflects the cost of definite states: decoherence destroys the correlations that permit maximal bi-predictability.

### 8.3 When can Bi-predictability Approach its Classical Ceiling?

The classical ceiling $P = 1/2$ is achieved only in an idealized limit in which the interaction is maximally information-preserving at the chosen description level. Two conditions are central:

- Determinism: $H(S' \mid S, A) = 0$. Given the current state and action, the next state is fully determined (vanishing forward predictive uncertainty $H_f$).
- Invertibility: $H(S, A \mid S') = 0$. The observed outcome uniquely identifies the state–action pair that produced it (vanishing backward predictive uncertainty $H_b$).

When both hold, forward and backward uncertainty are balanced and predictive asymmetry vanishes ($\Delta H = H_f - H_b = 0$). Passive physical systems can approximate this regime under sufficiently complete state descriptions, whereas agentic systems typically cannot: internal degrees of freedom, stochasticity, and many-to-one action–outcome mappings introduce residual forward and/or backward uncertainty, driving $P < 1/2$ and $\Delta H \neq 0$.

Additional technical requirements for saturating the ceiling include non-redundant action–state coupling, input–output entropy balance, and sufficient state capacity to permit bijective mappings.

### 8.4 Scope

All results assume discrete variables with finite entropies; logarithms are base-2. The bound $P \leq 1/2$is stated for Shannon entropy and does not directly extend to differential entropy, which can be negative and is not invariant to reparameterization. For continuous systems, we therefore compute $P$ after discretization at a fixed resolution.

# 9 Method: Physical Ground Truth: Double Pendulum

## 9.1 System and simulations

We analyze the double pendulum as a deterministic physical system with no agency. Two independent batches of 300 simulations each were generated from randomized initial conditions, covering both symmetric (equal bob masses) and asymmetric (unequal bob masses) configurations. Equations of motion were integrated using the Dormand-Prince method (ode45 in MATLAB) with strict error tolerances ($RelTol = AbsTol = 10^{-9}$) and maximum step size of 1 $ms$ (Strogatz, 2018; Press et al., 2007). Energy conservation was monitored; relative drift $|\Delta E(t)/E(0)|$ remained below 0.05% for all valid runs. The state vector $(\theta_1, \theta_2, \omega_1, \omega_2)$ was sampled at 1 kHz for subsequent analysis.

## 9.2 State representation

System state was represented by the full phase-space vector $S = (\theta_1, \theta_2, \omega_1, \omega_2)$. Angular variables were treated as circular quantities, while angular velocities were normalized to ensure comparability across trajectories. This representation captures the complete deterministic dynamics of the system.

## 9.3 Entropy Calculations

To estimate information-theoretic quantities from continuous-valued time series, it is necessary to construct empirical probability distributions over system states and their transitions. We therefore discretize the continuous state space using binning and estimate probabilities within finite temporal windows, which enables consistent computation of entropies, bi-predictability, and directional predictive uncertainty from observed trajectories.

To assess robustness to representation and temporal resolution, predictability metrics were evaluated using two independent discretization and windowing configurations. In the first batch, state variables were discretized using finer binning and longer sliding windows, while the second batch employed coarser binning and shorter windows with increased overlap. This design allows validation of the theory across distinct but reasonable choices of discretization and temporal aggregation, without altering the underlying dynamics.

## 9.4 Predictability Estimation

Bi-predictability ($P$), forward predictive uncertainty ($H_f$), backward predictive uncertainty ( $H_b$), and predictive asymmetry ($\Delta H$) were estimated from discretized state transitions using overlapping sliding windows. Probabilities were computed empirically within each window and aggregated across trajectories.

## 9.5 Chaos Quantification

Chaoticity was quantified using the finite-time Lyapunov exponent (FTLE), computed for each trajectory using standard methods. FTLE values were associated with corresponding $P$ estimates to assess the relationship between chaotic dynamics and Predictive Coherence. Across 600 simulations, $P$ showed no degradation with increasing FTLE (Table 8, Fig. 5), confirming robustness to chaotic dynamics. Observation: Chaotic trajectories exhibit a positive association with $P$, potentially reflecting improved entropy estimation from broader state-space coverage.

***Table 8*** *Chaos and bi-predictability. Summary statistics of finite-time Lyapunov exponents (FTLE) and bi-predictability (P) across both simulation batches.*

| Batch | Metric | Mean | STD |
|---|---|---|---|
| 1st batch | FTLE | -6.632019915 | 4.442605378 |
| | Bi-predictability P | 0.475747126 | 0.001552642 |
| 2nd batch | FTLE | -8.550904894 | 5.819810661 |
| | Bi-predictability P | 0.472919123 | 0.001692086 |

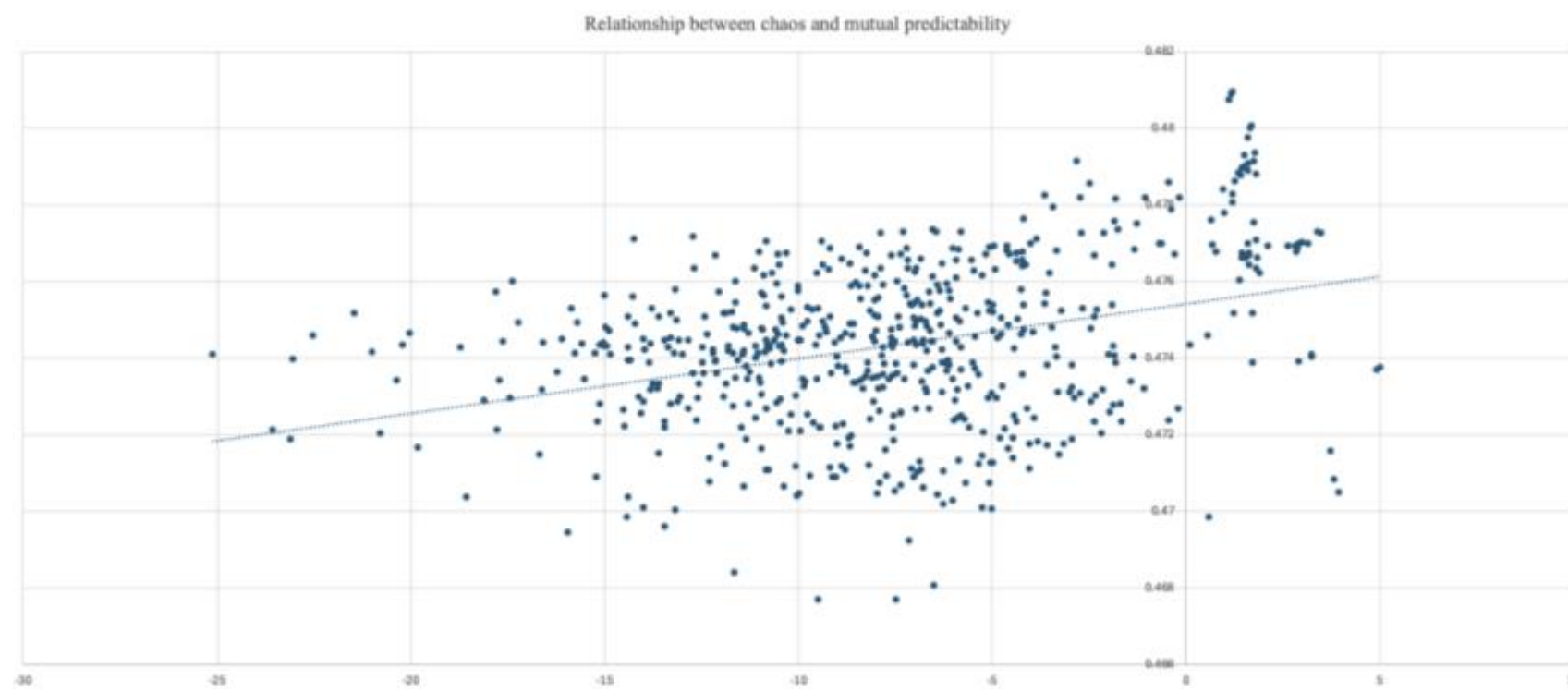

*__Figure 5__ Chaos correlates with bi-predictability in the double pendulum. Finite-time Lyapunov exponent (FTLE, x-axis) versus P (y-axis) across 600 simulations. The linear fit confirms a positive association; higher chaos corresponds to higher P, likely due to improved state-space coverage in entropy estimation. FTLE = finite-time Lyapunov exponent; P = bi-predictability.*

### 8.3 Discretization and Windowing

Continuous state variables were standardized (z-scored) and discretized into 16 equal-width bins per dimension. To characterize information dynamics over time, we employed sliding window analysis on the discretized trajectories. Each window spanned W = 300 time steps with a stride of $\delta = 75$ steps (75% overlap). Within each window, probability distributions $p(s)$ and joint distributions $p(s, s')$ were estimated using maximum likelihood from empirical state frequencies. Bi-predictability ($P$), forward and backward predictive uncertainties ($H_f$), ( $H_b$), and predictive asymmetry ($\Delta H$) were computed for each window.

### 9.6 Implementation

Double pendulum simulations were generated in MATLAB. All data processing, information-theoretic analysis, and statistical aggregation were performed on cloud-based compute infrastructure (Google Cloud Platform) to support large-scale analysis across hundreds of trajectories.

## 10 Method: RL Experiment Protocol

### 10.1 Reinforcement Learning Setup

Agents were trained and evaluated on the Half-Cheetah-v4 environment from MuJoCo (Todorov et al., 2012). The environment provides a standard continuous control benchmark with 17-dimensional state space (joint positions and velocities) and 6-dimensional action space (torque controls). Full environment specifications are available at [gymnasium.farama.org]. Agents were trained using Soft Actor-Critic (SAC) (Haarnoja et al., 2018) and Proximal Policy Optimization (PPO) (Schulman et al., 2017) with default hyperparameters from Stable-Baselines3 (Raffin et al., 2021). SAC agents were trained for approximately 2 million steps; PPO agents for approximately 1.5 million steps. Trained policies were frozen during evaluation. A total of 21 seeds were evaluated (11 SAC, 10 PPO). Five seeds were excluded from detection analyses because their pre-perturbation returns showed unstable baseline variance, making threshold calibration unreliable under the predefined protocol.

### 10.2 Perturbation Protocol

Each evaluation comprised 50 episodes (50,000 steps). $P, H_f, H_b,$ and $\Delta H$ were computed over sliding windows of 300 steps with stride 50, yielding 991 windows per run. Perturbations were injected at episode 15; preceding windows established baseline statistics for threshold computation. Eight perturbations were tested, spanning agent-side degradation and environment-side variability (Table 10):

***Table 9*** *RL perturbations settings. Eight perturbations spanning agent-side degradation (actuator noise, observation noise) and environment-side variability (external force, gravity).*

| Type | Category | Perturbation | Parameter |
|---|---|---|---|
| Agent | Actuator noise | act_noise_01% | 1% Gaussian noise on actions |
| Agent | Actuator noise | act_noise_03% | 3% Gaussian noise on actions |
| Agent | Actuator noise | act_noise_04% | 4% Gaussian noise on actions |
| Environment | External force | force_torso_5N | 5N force applied to torso, x- axis |
| Environment | External force | force_torso_10N | 10N force applied to torso, x-axis |
| Environment | Gravity | gravity_110% | Gravity increased to 110% |
| Agent | Observation noise | obs_noise_01% | 1% Gaussian noise on observations |
| Agent | Observation noise | obs_noise_03% | 3% Gaussian noise on observations |

### 10.3 Information Metric Computation

Computing information metrics on high-dimensional continuous spaces presents a tractability challenge. The Half-Cheetah environment yields 40 continuous variables per timestep (17 state + 6 action + 17 next-state). Direct joint distribution estimation over this space is infeasible.

We address this through structured dimensionality reduction:

1. Normalization: Each variable is z-scored independently
2. Discretization: Variables are binned into 3 equal-width bins
3. **Semantic grouping:** Variables are grouped by body part (front leg, back leg, torso) and concatenated to form composite symbols for $S, A$, and $S'$

This approach preserves the agent's embodiment structure while reducing continuous high-dimensional space to tractable discrete distributions. The choice of 3 bins was empirically determined; 4 bins yielded unreliable entropy estimates, and quantile-based binning produced flat, uninformative metrics. With 300 samples per window, strides 50, joint distribution estimation over composite symbols remains reliable. $P, H_f, H_b$, and $\Delta H$ were computed using standard entropy estimators with base-2 logarithms.

### 10.4 Detection Threshold and Latency

Detection thresholds were established using a $3\sigma$ criterion. For each metric ($P, H_f, H_b, \Delta H$, Reward), we computed the mean and standard deviation from baseline windows (episodes 1–14, prior to perturbation onset). A perturbation was considered detected when the metric exceeded 3 standard deviations from baseline mean. Detection latency was measured as the number of windows between perturbation onset (episode 15) and first threshold crossing. A latency of 0 indicates immediate detection in the first post-perturbation window. NaN indicates the perturbation was never detected.

IDT detection was defined as any component ($P, H_f, H_b, or\ \Delta H$) exceeding threshold. Detection rates were computed as the proportion of perturbation trials detected per seed, then averaged across seeds. IDT and Reward detection rates were compared using a paired t-test (n=21 seeds). Effect sizes were computed as Cohen's $d = (\mu_post - \mu_pre) / \sigma_pre$.

### 10.5 IDT Ensemble Detection

IDT detection was defined as any component ($P, H_f, H_b, or\ \Delta H$) exceeding threshold. This ensemble approach exploits component complementarity: $P\ and\ \Delta H$ respond independently to different perturbation types, and their union achieves higher coverage than any single metric.

## 11 Method: LLM Experimental Protocol

### 11.1 LLM Setup

A student model (Llama 3.1 8B, running on Ollama) engaged in multi-turn conversations with three teacher models: Claude Sonnet 4 (Anthropic, 2025), GPT-4o-mini (OpenAI, 2024), and Gemini Pro Preview (Google DeepMind, 2025). We conducted nine experimental tests organized into two categories:

Baseline bi-predictability tests examined natural conversation dynamics. Test 1 (Natural Variation, 85–150 turns) and Test 2 (Extended Variation, 100 turns) used graduated prompts progressing from simple questions to abstraction and revision. Test 3 (Semi-Random Variation, 200 turns) required unpredictable alternation across questioning styles. Test 4 (Long Memory, 200 turns) explored a single topic over a 32,768-token context window. Test 6 (Constrained Baseline, 150 turns) ran under constrained generation with Gemini only. Test 8 (Natural Dialogue Progression, 150 turns) combined elements of Tests 1–3 as a baseline for perturbation comparison.

Perturbation tests evaluated $P$'s sensitivity to conversational disruptions. Test 7 injected contradictions ("That doesn't sound right..."), Test 9 injected topic shifts ("Let's switch to discussing..."), and Test 10 injected non-sequiturs ("I had a sandwich yesterday..."). All perturbation tests followed identical structure: 30-turn baseline followed by five injections at turns 31, 46, 61, 76, and 91, with matched injection lengths (~40 words) to control for token count effects.

Tests were run under normal (temperature 0.7) and constrained (temperature 0.1) generation settings. In total, we collected approximately 4,500 conversation turns across 34 unique test-teacher-condition combinations. Student model parameters: 4096-token context limit, $top_p$ 0.9, $top_k$ 40, max response 150 tokens, repeat penalty 1.1.

### 11.2 Variable Mapping

The interaction stream was mapped to $(S, A, S')$ as follows:

- S: Accumulated context (all prior tokens, grows each turn)
- A: Student response (current turn tokens only)
- S': Teacher prompt (current turn tokens only)

Token distributions were computed using the Llama-2-7b-hf tokenizer (NousResearch). To establish baseline bi-predictability metrics, we employed two widely-adopted approaches. First, semantic similarity between prompt-response pairs was computed using cosine similarity of sentence embeddings generated by Sentence-BERT (Reimers & Gurevych, 2019), specifically the all-MiniLM-L6-v2 model. Second, response quality was assessed using the LLM-as-a-Judge paradigm (Zheng et al., 2023), which has demonstrated approximately 80% agreement with human evaluators. GPT-4 was used to score each response on a scale of 1-7 for relevance, coherence, and helpfulness.

### 11.3 Information Metric Computation

For each turn, entropy and information metrics were computed from token frequency distributions:

***Table 10*** *Metrics tracked during the LLM simulation*

| Metric | | Formula |
|---|---|---|
| States Entropy | $H(S)$ | Shannon entropy of accumulated context |
| Actions Entropy | $H(A)$ | Shannon entropy of student response |
| Next States Entropy | $H(S')$ | Shannon entropy of teacher prompt |
| States-Actions Mutual Information | $MI(S; A)$ | $H(S) + H(A) - H(S, A)$ |
| Bi-predictability | $P$ | $MI(S, A; S') / [H(S) + H(A) + H(S')]$ |
| Forward Predictive Uncertainty | $H_f$ | $H(S, A, S') - H(S, A)$ |
| Backward Uncertainty | $H_b$ | $H_b$ |
| Predictive Asymmetry | $\Delta H$ | $H_f$ - $H_b$ |

All entropy calculations used base-2 logarithms.

## 11.4 Perturbation Protocol

Perturbations were injected at turns 31, 46, 61, 76, and 91. Injections consisted of contradiction or confusion statements designed to disrupt conversational coherence. Turns 1–30 established baseline statistics.

## 11.5 Validation Metrics

To compare P against semantic evaluation methods, we computed:

- cosine_sim: SentenceTransformer (all-MiniLM-L6-v2)
- adjacent_coherence: Cosine similarity between consecutive responses
- cumulative_drift: Cosine similarity between response and first response
- LLM-judge: MT-Bench style scoring (GPT-4o-mini)

Detection was compared across $P$, $H_b$ cosine_sim, and LLM-judge at each injection point. Experiments were conducted on Azure (Standard_NC4as_T4_v3, NVIDIA T4 GPU) with API access to Claude, GPT-4o-mini, and Gemini teacher models. Evaluation comprised over 4,000 conversational turns across three teacher models and multiple experimental conditions, providing robust statistics for detection comparison.

Effect sizes were large across all metrics (Cohen's $d > 0.8$ in all 36 comparisons). $P$ showed effect sizes ranging from $d = 1.26$ $to$ $d = 6.99$, comparable to LLM-as-Judge ($d = 2.22 - 4.55$). Correlations between P and cosine similarity were significant with narrow confidence intervals (e.g., Test 7/Claude: $r = 0.838, 95\%$ $CI$ $[0.77, 0.89]$).